\documentclass[10pt,twocolumn,letterpaper]{article}

\usepackage[pagenumbers]{iccv} % To force page numbers, e.g. for an arXiv version

\definecolor{iccvblue}{rgb}{0.21,0.49,0.74}
\usepackage[pagebackref,breaklinks,colorlinks,allcolors=iccvblue]{hyperref}
\usepackage[table]{xcolor}
\usepackage{stfloats}
\usepackage{algorithmic}
\usepackage{algorithm}
\usepackage{pifont}
\newcommand{\xmark}{\ding{55}}
\usepackage{multirow}
\def\paperID{7463} % *** Enter the Paper ID here
\def\confName{ICCV}
\def\confYear{2025}
\newcommand{\incre}[1]{\textcolor{green!60!black}{\small$\uparrow$ #1}}
\title{MTC-VAE: Multi-Level Temporal Compression with Content Awareness}

\author{
  \textbf{Yubo Dong\textsuperscript{1}},
  \textbf{Linchao Zhu\textsuperscript{1,$\dag$}}
\\
  \textsuperscript{1}ReLER, CCAI, Zhejiang University
\\
  \textsuperscript{$\dag$}Corresponding author
}

\begin{document}
\maketitle
\begin{abstract}
{L}atent {V}ideo {D}iffusion {M}odels (LVDMs) rely on Variational Autoencoders (VAEs) to compress videos into compact latent representations. 
For continuous Variational Autoencoders (VAEs), achieving higher compression rates is desirable; yet, the efficiency notably declines when extra sampling layers are added without expanding the dimensions of hidden channels. In this paper, we present a technique to convert fixed compression rate VAEs into models that support multi-level temporal compression, providing a straightforward and minimal fine-tuning approach to counteract performance decline at elevated compression rates.Moreover, we examine how varying compression levels impact model performance over video segments with diverse characteristics, offering empirical evidence on the effectiveness of our proposed approach. We also investigate the integration of our multi-level temporal compression VAE with diffusion-based generative models, DiT, highlighting successful concurrent training and compatibility within these frameworks. This investigation illustrates the potential uses of multi-level temporal compression.
\end{abstract}

\section{Introduction}
\label{sec:1}
% 第一段介绍都有哪些VAE 视频生成有多火，两套视频生成范式。
The field of video generation has attracted considerable interest from both academia and industry, notably following OpenAI's announcement of SORA \cite{videoworldsimulators2024}. Presently, Latent Video Diffusion Models (LVDMs) like MagicTime \cite{magictime}, VideoComposer \cite{wang2024videocomposer}, AnimateDiff \cite{guoanimatediff}, Stable Video Diffusion (SVD) \cite{blattmann2023stable}, HiGen \cite{qing2024hierarchical}, Latte \cite{ma2024latte}, SORA \cite{videoworldsimulators2024}, Open-Sora \cite{opensora}, Open-Sora-Plan \cite{pku_yuan_lab_and_tuzhan_ai_etc_2024_10948109}, CogVideo \cite{hong2022cogvideo}, CogVideoX \cite{yang2024cogvideox} and Hunyuan-DiT
\cite{li2024hunyuanditpowerfulmultiresolutiondiffusion}
have come to lead the video generation space due to their reliability, performance, and adaptability. The common workflow among these LVDMs involves using Variational Autoencoders (VAEs) \cite{kingma2014auto} to compress original videos into latent variables. Subsequently, denoisers are trained to estimate the noise applied to these compacted forms. The Stable Diffusion VAE (SD-VAE) \cite{rombach2022high, podell2023sdxl}, employed by LVDMs, focuses on compressing individual image frames without accounting for temporal consistency, which leads to increased input sizes and hardware needs while diminishing the quality and accuracy of reconstruction. Prior studies have investigated the use of VQ-VAEs \cite{van2017neural} for temporal compression in video generation by transforming videos into discrete tokens for autoregressive models \cite{yan2021videogpt, ge2022long, yu2023magvit, yu2023language}.

\vspace{0.25cm} 
% 逻辑 讲它很重要，现有的工作主要做法是什么，训练和生成的过程讲清楚
\noindent VideoVAE is a critical component for LVDM. Previous approaches, like Stable Video Diffusion VAE (SVD-VAE) \cite{blattmann2023stable}, OD-VAE \cite{pku_yuan_lab_and_tuzhan_ai_etc_2024_10948109}, and CV-VAE \cite{zhao2024cv}, integrate temporal data and prioritize enhancing quality with predetermined compression rates. These methods encode video segments into latents simultaneously for greater efficiency and later combine them for diffusion training. In the inference phase, a diffusion transformer generates the latent video from text prompts. It then divides this latent video into segments, which are individually decoded into videos using the VAE decoder. These segments are subsequently merged. Due to varying video dynamics, the predefined temporal compression rates may not suit all video segments.

\vspace{0.25cm}
% 我的设计就是为了探索能不能实现一种多级时间压缩，在不影响能力的前提下提升压缩率。为此我展开了这个工作
\noindent To investigate the potential of enhancing temporal compression rate without compromising reconstruction quality, we introduce the \textbf{MTC-VAE}, or \textbf{M}ulti-level \textbf{T}emporal \textbf{C}ompression VAE, which is designed to be seamlessly integrated with pretrained VAEs. In contrast to VAEs with a fixed temporal compression rate, \textbf{MTC-VAE} compresses video segments temporally into latent representations by accounting for the distinctive attributes of each segment. It is compatible with a wide range of 3D-VAEs as foundational models and adapts video reconstruction accordingly. Drawing on the notion that segments which are sparse or exhibit slow movement can undergo higher compression compared to those that are dense or rapidly changing, we propose an approach to dynamically assign compression rates to video segments based on their temporal complexity. However, this approach presents a technical difficulty: since a diffusion transformer is trained using latent videos generated from combined latent segments, how can these be accurately decoded back to their initial temporal resolution? We formulated the Video Clipper module, adept at pinpointing the initial frame (keyframe) of each segment within the hidden features, facilitating accurate decompression of each segment conforming to its initial compression rate. Additionally, we propose a flow-guided loss to adaptively optimize the balance between compression efficiency and reconstruction quality across various video dynamics (Table~\ref{Table1}).

% 这一段接着讲我们的方法实现的VAE能不能支持DiT的训练和视频生成，结论是可以。
\vspace{0.25cm}
\noindent Can our MTC-VAE be seamlessly combined with a pre-trained DiT model? To explore this, our MTC-VAE is utilized to fine-tune DiT. We carried out experiments where the MTC-VAE Encoder was employed to transform videos into latent representations at different temporal compression rates. These representations were linked with the corresponding video text descriptions to serve as finetuning inputs for the DiT model. We adjusted two open-source Latent Video Diffusion Models (LVDMs) and observed that these models could effectively accommodate the fluctuating temporal compression features of our MTC-VAE. This adaptation could facilitate the generation of longer videos with the same hardware or lower the computational demands for videos of similar duration, optimizing the balance between quality and compression efficiency (Table~\ref{tab:vbench}).

\vspace{0.25cm}
\noindent Our MTC-VAE experiments highlight that we can significantly boost the compression rate by 92.4\% while still achieving nearly identical performance, with only marginal differences of 0.0027 in SSIM and 0.15 in PSNR. Our detailed analysis, presented as scatter plots comparing quality metrics to compression rates across various video genres, uncovers an interesting insight: lower temporal compression rates in VAEs do not consistently lead to the best outcomes across all video styles. This complex relationship between compression rate and quality supports the core idea of our adaptive strategy, illustrating how selectively applying elevated compression rates to suitable video segments preserves quality and reduces overall compression rates.

\begin{figure*}[h]
\centering
\includegraphics[width=1.0\textwidth]{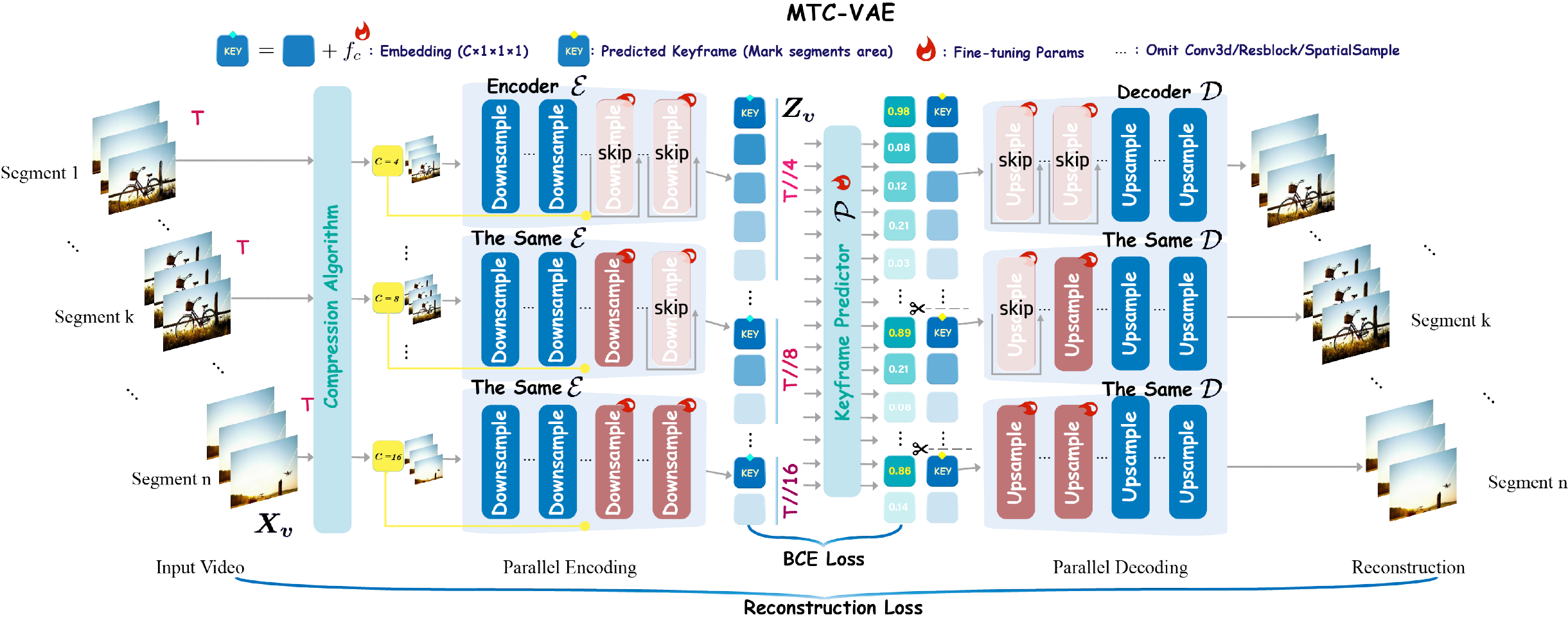}
\caption{Our MTC-VAE model is illustrated. This architecture incorporates a compression algorithm that adaptively manages \textbf{temporal compression rates} \(C\) across video segments, deciding the quantity of sampling layers to implement. Only the temporal sample layers are shown, with newly introduced sample layers initialized from existing pretrained ones. The segment area is indicated by the embedding \(f_c\). Our linear keyframe predictor is adept at identifying the labeled latent features from the combined latent representation, and segments are divided based on these features. These latent segments are then decoded to generate the final video output. The specific process can be referenced in the pseudocode in Algorithm~\ref{alg:vae}.}
\label{fig:flex_vae}
\end{figure*}

\section{Related Work}
\label{sec:2}

\subsection{Latent Video Diffusion Model }
\label{sec:2.1}
LVDMs represent a crucial challenge in artificial intelligence \cite{Prompt2Poster}. They employ VAEs to encapsulate video content into latent formats, followed by leveraging denoisers for noise prediction, a field experiencing swift advancements. OpenAI's SORA \cite{videoworldsimulators2024}, which creates 1080P resolution videos lasting one minute, significantly amazes the world. LVDMs are categorized based on their denoiser frameworks. The first category incorporates U-net-based denoisers \cite{ronneberger2015u, ChronoMagic-Bench, clh2023end}, like MagicTime \cite{magictime}, AnimateDiff \cite{guoanimatediff}, and Stable Video Diffusion (SVD) \cite{blattmann2023stable}. Meanwhile, the second category employs Transformer-based denoisers \cite{peebles2023scalable}, including Latte \cite{ma2024latte}, SORA \cite{videoworldsimulators2024}, CogVideoX \cite{yang2024cogvideox}, Vidu \cite{bao2024vidu}, and Hunyuan-DiT~\cite{li2024hunyuanditpowerfulmultiresolutiondiffusion}. Regardless of the denoiser structures, VAEs dictate the input dimensions for denoisers and the fidelity of video reconstructions from latent forms. Therefore, VAEs offering efficient representation while ensuring high-quality reconstruction will substantially boost performance.

% 简单的重新讲一下现有的VideoVAE并指明我们要进行比较。
\subsection{Variational Autoencoder }
\label{sec:2.2}
VAE, originally crafted to optimize the Evidence Lower Bound (ELBO) for generative tasks \cite{kingma2014auto}, is now integral to various generative models, split into two classifications. First, VQ-VAEs \cite{van2017neural}, which convert videos into discrete tokens for use with autoregressive models \cite{ge2022long, yu2023magvit, yu2023language}. MAGVIT-v2 \cite{yu2023language} excels in video reconstruction yet its discrete tokens don't fit LVDMs. Conversely, continuous VAEs create continuous video representations for LVDMs; notable ones include SD-VAE \cite{rombach2022high} and SVD-VAE \cite{blattmann2023stable}. These, however, ignore temporal redundancy by focusing solely on spatial compression. Recent studies are trialing static temporal compression rates. OPS-VAE \cite{opensora} employs two VAEs for spatial and temporal compression, CV-VAE \cite{zhao2024cv} uses temporal compression to align with SD-VAE, and OD-VAE \cite{chen2024odvaeomnidimensionalvideocompressor} implements a 3D-causal-CNN for combined compression. We compared our MTC-VAE against these methods in experiments.

% overview 讲一下每一个sec在讲什么
\section{Method}
\label{sec:3}
In this work, we present the \textbf{MTC-VAE} (\S\ref{sec:3.1}), short for \textbf{M}ulti-level \textbf{T}emporal \textbf{C}ompression VAE, designed to seamlessly integrate with existing pretrained VAEs. We describe a core compression algorithm (\S\ref{sec:3.2}) where parameters are set heuristically, following our analysis that suggests VAE compression quality is affected by the properties of the video segments. Subsequently, we discuss the Encoder procedure (\S\ref{sec:3.3}). To restore segments to their original temporal resolution, we developed a video clipper (\S\ref{sec:3.4}) and tested various configurations, considering that a diffusion transformer is trained on latent videos constructed from combined segments. Additionally, we elaborate on the fine-tuning strategies (\S\ref{sec:3.5}) and essential structural alterations necessary to adopt our proposed method in this section.
% 讲一下视频在VAE中的编解码过程。
\subsection{MTC-VAE}
\label{sec:3.1}
Our MTC-VAE consists of an encoder \(\mathcal{E}\) and a decoder \(\mathcal{D}\). As depicted in Figure~\ref{fig:flex_vae}, the video is initially segmented into fixed-length video segments with a resolution of T×H×W. The multi-level compression algorithm determines a temporal compression rate \(C\) by balancing quality and compression. The encoder then converts the video segment into a latent segment with dimensions t×h×w, determining the number of temporal sample layers based on the compression rate \(C\). Each segment receives an additional embedding to indicate the segment area. During decoding, a linear video predictor \(\mathcal{P}\) is utilized to identify labeled features and divide the latent representation into segments. We use the latent length and fixed video length to calculate the temporal compression rate and decode it back. Ratios such as 4×8×8 imply that \(\frac{T}{t}=4,\frac{H}{h}=8,\frac{W}{w}=8\), among others. New up/down sampling layers (incorporating nearest-neighbor sampling layers and weights from the base video VAE) are integrated into the original architecture to achieve compression ratios of 4×8×8, 8×8×8, and 16×8×8.

%讲清楚公式的设计为什么是这样的
\subsection{Compression Algorithm}
\label{sec:3.2}
To enhance temporal compression while maintaining comparable reconstruction efficacy, the main challenge our algorithm addresses is finding an optimal balance between reconstruction \textbf{quality} and \textbf{compression} efficiency. We tackle this by optimizing a score function for the input video. Consider a video divided into \(N\) segments, represented by the tensor \(\boldsymbol{X} = [\boldsymbol{x}_1, \boldsymbol{x}_2, \ldots, \boldsymbol{x}_N] \in \mathbb{R}^{N \times T \times H \times W \times 3}\), where each segment \(\boldsymbol{x}_i\) includes \(T\) frames. Our objective is to maximize the total score:
\begin{align}
\sum_{i=1}^N \Big[&(1-\alpha(x_i)) \cdot Q(c_i, x_i) + \alpha(x_i) \cdot \log_2(c_i) \cdot w\Big],\\
\text{s.t. }& c_i\in \{4,8,16\}, \forall i \in \{1,2,...,N\}.
\end{align}
This summation is made up of two elements: the initial segment evaluates the \textbf{quality} of each part at designated compression levels, whereas the latter segment encourages increased \textbf{compression} rates, with \(w\) managing the tendency towards more compression, with a default of \(w = 1.5\).

\noindent The quality assessments produce a quality vector \(\boldsymbol{Q}(x_i) = [Q(4,x_i), Q(8,x_i), Q(16,x_i)]\) for each segment. The compression tolerance factor \(\alpha(x_i)\) for a given segment is computed as follows:
\begin{equation}
\alpha(x_i) = \frac{1}{2} \cdot \frac{\overline{Q}(x_i) - Q_{\text{min}}}{Q_{\text{max}} - Q_{\text{min}}} + \frac{1}{2} \cdot \left(1 - \frac{\sigma_Q(x_i)}{\sigma_{Q_\text{max}}}\right).
\end{equation}
Where:
\(\overline{Q}(x_i)\) signifies the average quality over various compression rates for segment \(x_i\),
\(Q_{\text{min}}\) and \(Q_{\text{max}}\) denote the smallest and largest PSNR values across all compression rates and segments,
\(\sigma_Q(x_i)\) represents the standard deviation of quality across compression rates for segment \(x_i\),
\(\sigma_{Q_\text{max}}\) is the highest standard deviation among all segments. Utilizing PSNR to evaluate \(Q(·)\) proves more effective, as demonstrated in Table \ref{tab:cmp_Q}. This model increases \(\alpha(x_i)\) (promoting greater compression) when:
1. The segment achieves higher average quality across compression rates.
2. The segment exhibits minimal quality variation across different compression rates.

\begin{figure*}[t]
    \centering
    \includegraphics[width=1.05\linewidth]{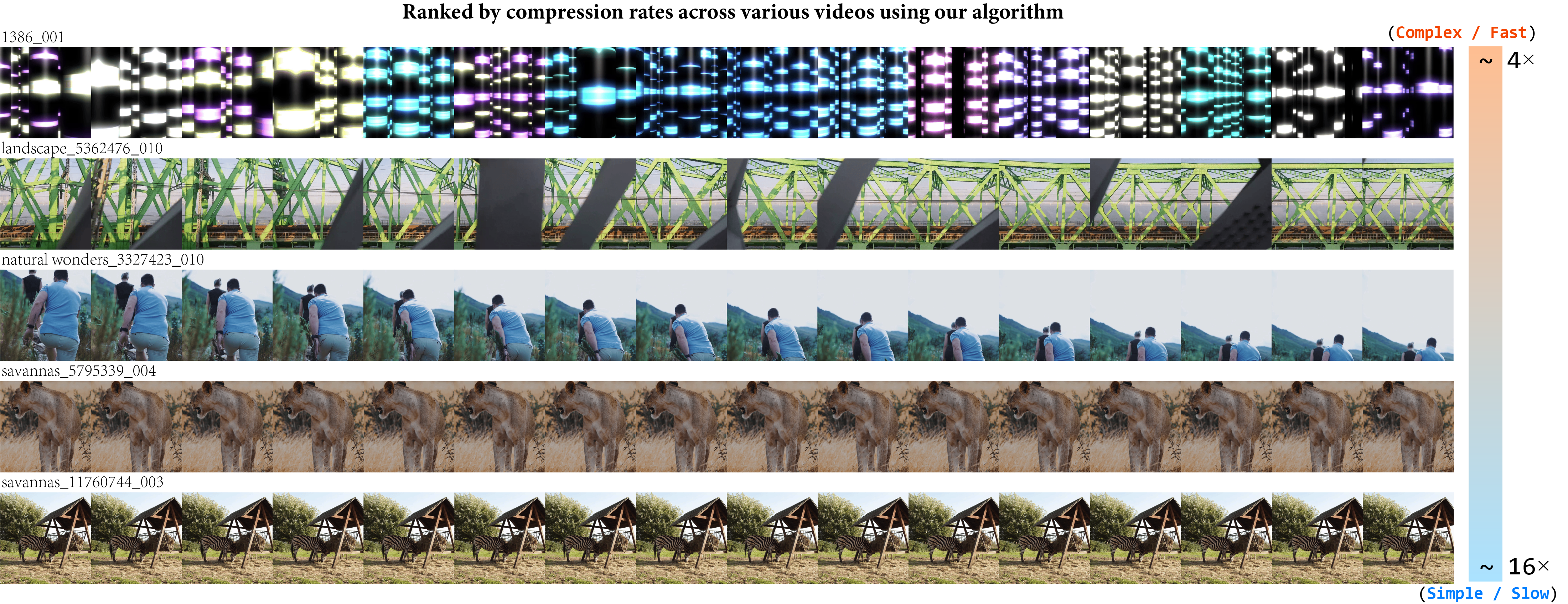}
    \caption{Using our compression algorithm, we clearly observe that it tends to apply higher compression rates to simpler and slower videos, while employing lower temporal compression rates for more complex or fast-moving videos.}
    \label{fig:scatter}
\end{figure*}

\subsection{VAE Encoder}
\label{sec:3.3}
Encode the video through a three-step process: Initially, determine the suitable compression rates for each video segment using our algorithm. Next, compress each segment in parallel and incorporate a keyframe embedding for each. Finally, concatenate them along the temporal dimension.

\noindent\textbf{Encoding Procedure.}  
Consider the video segments \(\boldsymbol{X}\). The compressed latent representation is expressed as \(\boldsymbol{Z} \in \mathcal{R}^{N \times t \times h \times w \times c}\). Initially, MTC-VAE employs our compression algorithm~\ref{sec:3.2} to determine a sequence of compression rates \(\boldsymbol{C}\), defined as \(\boldsymbol{C} = [{c_1}, {c_2}, \ldots, {c_N}] \in \mathcal{R}^N\), where the compression rate for the \(i\)-th video segment is indicated by \({c_i}\). Subsequently, the video segments \(\boldsymbol{X}\) are processed in parallel by \(\mathcal{E}\), potentially omitting sample layers in accordance with the compression rates in \(\boldsymbol{C}\) to derive the latent features:
\begin{equation}
\boldsymbol{z}_i = \mathcal{E}(\boldsymbol{x}_i, c_i), \quad i \in \{1,\ldots,N\}.
\end{equation}For the first latent feature \(\boldsymbol{z_i}\) at \(t=0\) of each segment, known as the \textbf{keyframe}, an extra embedding \(f_c\) is included, see Figure~\ref{fig:flex_vae}. Integration of \(f_c\) at the final frame or its exclusion is detailed in table \ref{extra_emb}):
\begin{equation}
\boldsymbol{z_i}[:,j] = 
\begin{cases} 
\boldsymbol{z_i}[:,j] + f_c, & \text{if } j = 0 \\ 
\boldsymbol{z_i}[:,j], & \text{otherwise} 
\end{cases}
\end{equation}
The ultimate latent representation $\boldsymbol{Z}$ is formed by concatenating these individual segment encodings:
\begin{equation}
\boldsymbol{Z} = [\boldsymbol{z}_1 \| \boldsymbol{z}_2 \| \ldots \| \boldsymbol{z}_N], \quad \boldsymbol{Z}\in \mathcal{R}^{Nt \times h \times w \times c}.
\end{equation}

\subsection{VAE Decoder}
\label{sec:3.4}
Given that segments are compressed at various temporal rates, the initial task is to retrieve the specific temporal compression rates for the complete latent feature. Upon ascertaining the temporal compression rates from the latent feature, each segment can be concurrently decoded based on its particular temporal compression rate.
\noindent\textbf{Recover Temporal Compression Rates.}
Extracting temporal compression rates from latent features poses a challenge. We propose two structures to tackle this issue:
a) \textbf{Analysis per latent unit}. We utilize a predictor to assess the temporal compression rate at each temporal index within the latent segment and leverage the smoothed compression rate outcomes at each location to approximate the actual segment compression rate. However, this method requires training a greater number of parameters, which can substantially diminish reconstruction capabilities.
b) \textbf{Convert multi-classification to binary classification}. In response to the complexity of the previous predictor, we develop a novel keyframe predictor. Its sole function is to identify keyframes (the initial frame of each latent segment). The keyframe predictor
\(\mathcal{P}\) serves as a binary classifier that moves through \(\boldsymbol{Z}\) over time to generate logits to detect latent keyframe latents, used to compute \(\boldsymbol{\hat{C}}\) from the fixed length segment $T$. Then, \(\boldsymbol{Z}\) is divided temporally at these identified keyframe latents, which can be formatted as:
\begin{equation}
\boldsymbol{\hat{C}} = \left\{ \frac{T}{\text{length}(\boldsymbol{Z}[i])} \mid i \in N\right\}.
\end{equation}

\noindent\textbf{Decoding procedure.} 
Based on the keyframes identified by the predictor, latents are divdied into distinct latent segments. Each of these segments \(z_i\) is then decoded into video parts \(\hat{x}\) and subsequently combined:
 \begin{equation}
\hat{x}_i = \mathcal{D}({z}_i, \hat{c}_i), \quad i \in \{1,\ldots,N\}.
\end{equation}

\subsection{Training}
\label{sec:3.5}
We applied a two-step fine-tuning method. Initially, we quickly fine-tuned the VAE at compression rates of 8×8×8 and 16×8×8. By adding two sampling layers, based on the nearby sampling layer of the pre-trained VAE and initialized with its parameters, we adjusted these layers for their new roles. The first stage objective involves combining reconstruction, adversarial, and KL regularization losses, using the loss as same as OD-VAE~\cite{chen2024odvaeomnidimensionalvideocompressor}.
We used the Adam optimizer \cite{kingma2014adam} to perform 10k fine-tuning steps in stage 1 with a learning rate of \(1 \times 10^{-5}\), using 4 A100 GPUs. The VAE training dataset included open-source videos from OpenSoraPlan \cite{pku_yuan_lab_and_tuzhan_ai_etc_2024_10948109}. Input videos were divided into 17-frame segments at a \(256 \times 256\) pixel resolution.

% \vspace{0.25cm}
\noindent In the second phase, we trained the embedding \(f_c\) (c×1×1×1) and the linear Keyframe Predictor, utilizing extra losses: \textbf{Binary cross-entropy loss}. The keyframe detector \(\mathcal{P}(\cdot)\), identifying if a latent frame \(Z_i\) is a keyframe, relies on this loss:
\begin{equation}
\mathcal{L}_{\text{BCE}} = -\frac{1}{N} \sum_{i=1}^{N} \left[ y_i \log(p_i) + (1-y_i) \log(1-p_i) \right],
\end{equation}
with \(p_i = \mathcal{P}(Z_i)\) as the predicted keyframe probability, and \(y_i\) as the true label. \textbf{Flow-guided consistency loss.} We found that enhancing both motion and quality aspects improves performance. The flow-guided consistency loss is:
\begin{equation}
\mathcal{L}_{\text{flow}} = \mathcal{L}_{\text{quality}} + \mathcal{L}_{\text{motion}},
\end{equation}
defining quality and motion losses as:
\begin{equation}
\mathcal{L}_{\text{quality}} = \| \text{warp}(\hat{\boldsymbol{X}}, \Delta \mathbf{f}) - \boldsymbol{X} \|_1,
\end{equation}
\begin{equation}
\mathcal{L}_{\text{motion}} = \| \Delta \mathbf{f} \|_1.
\end{equation}
Here, \(\Delta \mathbf{f}\) is the optical flow from the optical flow model \cite{Morimitsu2024RecurrentPartialKernel}, and \(\text{warp}(\hat{\boldsymbol{X}}, \Delta \mathbf{f})\) is the adjusted output using this flow. This stage included 10k additional training steps.
To enhance stability, we implement an exponential moving average (EMA) during training with a decay rate of 0.999, following the approach proposed in \cite{saito2020train, tran2015learning}. 

\begin{table*}[htbp]
\footnotesize
 \centering
 \resizebox{\linewidth}{!}
 {\begin{tabular}{l|c|c|cccc|cccc}
 \hline
   \multirow{2}{*}{Model}& \multirow{2}{*}{Pattern}&\multirow{2}{*}{Ch}& \multicolumn{4}{c}{WebVid-10M} & \multicolumn{4}{|c}{Panda-70M}\\
   & &&VCPR $ \uparrow$ 
&PSNR $\uparrow$&SSIM $\uparrow$&LPIPS $\downarrow$
&VCPR $\uparrow$
&PSNR $\uparrow$&SSIM $\uparrow$&LPIPS $\downarrow$
\\
\hline
VQGAN~\cite{esser2021taming}&( 1 × 8 × 8 )& 4&64 
&26.26&0.7699&
 0.0906
&
64
&26.07&0.8295&0.0722
\\
SD-VAE~\cite{rombach2022high}&( 1 × 8 × 8 )&4&64 
&30.19&0.8379&0.0568
&
64
&30.40&0.8894&0.0396
\\
SVD-VAE~\cite{blattmann2023stable}&( 1 × 8 × 8 )&4&64 
&31.15&0.8686&
 0.0547
&
64
&31.00&0.9058&0.0379
\\
% \hline
TATS~\cite{ge2022long}&( 4 × 8 × 8 )&4&256 
&23.10&0.6758&0.2645
&
256
&21.77&0.6680&0.2858
\\
CV-VAE~\cite{zhao2024cv}&( 4 × 8 × 8 )&4&256 
&30.76&0.8566&
 0.0803
&
256
&29.57&0.8795&0.0673
\\
OPS-VAE~\cite{opensora}&( 4 × 8 × 8 )&4&256 
&31.12&0.8569&0.1003
&256
&31.06&0.8969&0.0666
\\
OD-VAE~\cite{pku_yuan_lab_and_tuzhan_ai_etc_2024_10948109}&( 4 × 8 × 8 )&4&256 
&31.16&0.8694&
 0.0586
&
256
&30.49&0.8970&0.0454
\\
CogVideoVAE~\cite{yang2024cogvideox}&( 4 × 8 × 8 )&16&256&\textbf{35.25}&\textbf{0.9323}&
 \underline{0.0317}&
256&\textbf{34.92}&\textbf{0.9422}&\underline{0.0206}\\
\hline
\rowcolor[gray]{0.9} MTC-VAE & ( / × 8 × 8 )& 4& \underline{443.7}\incre{+73.3\%}
& 31.25& 0.8703& 0.0621
& \underline{371.9}\incre{+45.3\%}& 30.69& 0.8993&0.0464
\\
\rowcolor[gray]{0.9} MTC-VAE &( / × 8 × 8 )&16&\textbf{492.5}\incre{+92.4\%} &\underline{35.10}&\underline{0.9296}&\textbf{0.0304}&\textbf{431.2}\incre{+68.4\%}&\underline{34.84}&\underline{0.9409}&\textbf{0.0205}\\
\hline
 \end{tabular}
 }
 \caption{Video reconstruction results of VAEs on WebVid-10M and Panda-70M validation sets, including compression rates (VCPR).Our MTC-VAE still performs comparably to the CogVideoVAE model with higher temporal compression rates.
 "Ch" means the latent feature channel size. / indicates multi-level temporal compression. Best and second-best scores are in \textbf{bold} and \underline{underlined}.}
\label{Table1} 
 \end{table*}

\begin{figure*}[ht]
\centering
\includegraphics[width=1.0\textwidth]{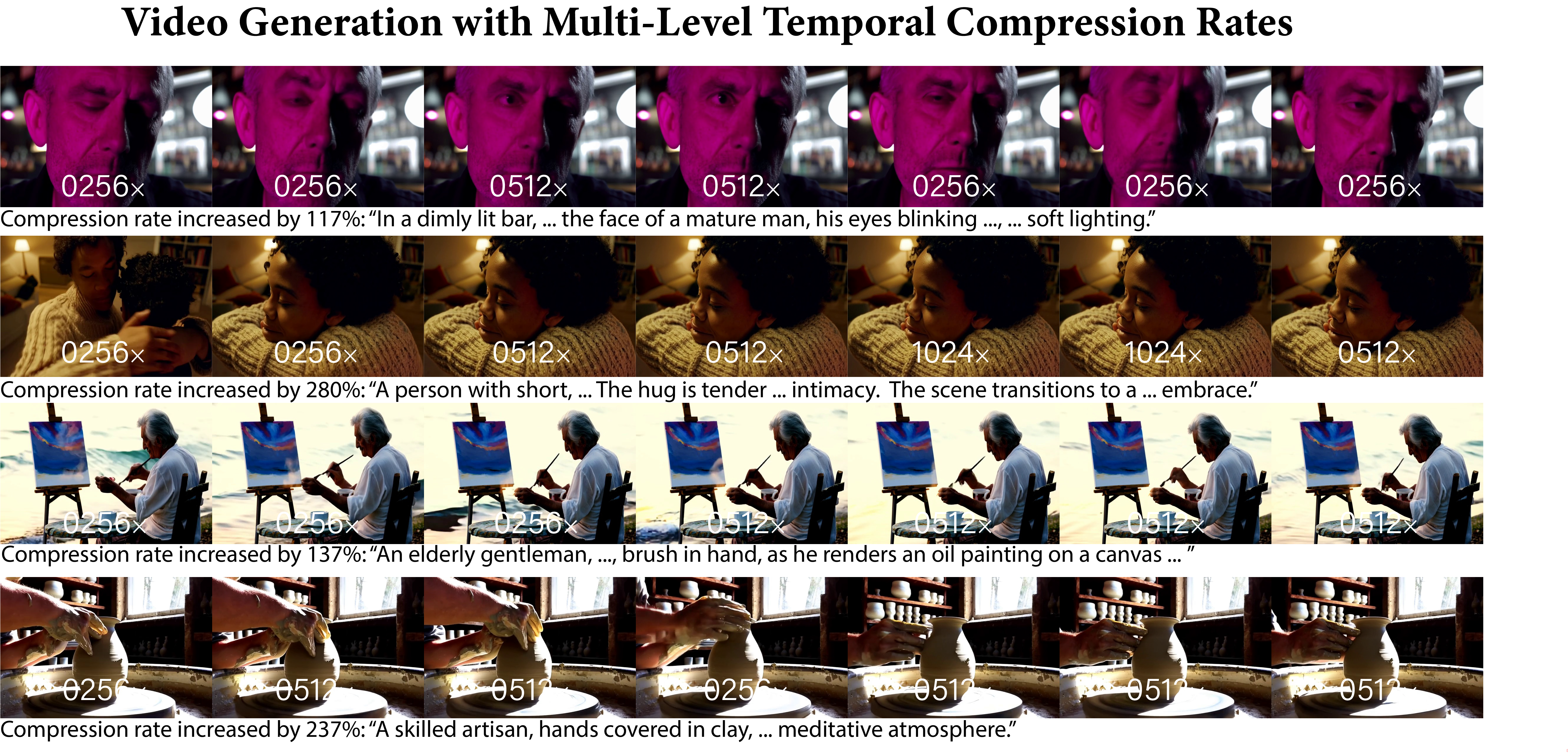}
\vspace{-2.0em}
\caption{MTC-VAE enables a finely-tuned DiT to allocate less latent space for \textbf{simple} data with \textbf{high} compression, while dedicating more for \textbf{complex} frames with \textbf{lower} compression rates. We illustrate four videos generated with differing compression: 256× (4×8×8), 512× (8×8×8), and 1024× (16×8×8). This method assists DiT in creating videos with variable frame lengths. The generation process includes producing 720p videos in 50 sampling steps, utilizing carefully designed prompts to emphasize temporal scene transitions. Additional visuals and comparisons can be found in the supplementary~\ref{fig:appearance} and material.
}
\label{fig2}
\end{figure*}

\begin{table*}[h]
\footnotesize
\centering
\resizebox{\linewidth}{!}{\begin{tabular}{l|c|cccclc}
\hline
\multirow{2}{*}{Model}& \multirow{2}{*}{Ch} & VCPR $ \uparrow$ & Latency $\downarrow$ & Speedup $\uparrow$  & VBench $\uparrow$ & \multirow{2}{*}{Resolution}& \multirow{2}{*}{Steps}\\
& & (times) & (sec/frame) & (times)  & Total(\%) & & \\
\hline
OpenSora-Plan~\cite{pku_yuan_lab_and_tuzhan_ai_etc_2024_10948109} &  4 &256.0 &  1.36
&$1.00\times$  &  74.34 
&512$\times$512& 30\\
\rowcolor[gray]{0.9} MTC-OSP&  4&\textbf{441.9}  &  \textbf{0.71}
&\textbf{1.91}$\times$  &  
74.72 
&512$\times$512& 30\\
\hline
CogVideoX-5B~\cite{yang2024cogvideox} &  16&256.0 &  4.43
&$1.00\times$  &  80.88 
&720$\times$480& 50\\
\rowcolor[gray]{0.9} MTC-CogVideo&  16&366.7  &  3.12
&$1.42\times$  &  80.70 
&720$\times$480& 50\\
\hline
\end{tabular}}
\caption{Evaluation results for video generation. This table compares VCPR, speedup, latency and sample steps for CogVideoX and OpenSora-Plan, both with and without our MTC-VAE. "Ch" means the latent feature channel size.}
\label{tab:vbench}
\end{table*}

\section{Experiment}
\label{sec:4}
Initially, we detail the experimental configuration and the evaluation metrics (\S\ref{sec:4.1}). Subsequently, we carry out a comprehensive comparison of MTC-VAE with various baseline models for video reconstruction. We explore the correlation between video features and tolerance compression rates (\S\ref{sec:4.2}). Furthermore, we conduct an analysis of the quantitative reconstruction results as well as the generation outcomes (\S\ref{sec:4.3}). Additionally, a series of ablation studies are performed (\S\ref{sec:4.4}).

\subsection{Experimental Setup}
\label{sec:4.1}
\textbf{Baselines.}
To evaluate the performance and efficiency, we compared our MTC-VAE against eight state-of-the-art VAEs for video reconstruction. The models include:
(1) VQGAN, SD-VAE, and SVD-VAE: These are image VAEs based on VQ techniques. VQGAN \cite{esser2021taming} is a popular image VQ-VAE (f8-8192 version) \cite{zhao2024cv}. SD-VAE \cite{rombach2022high} is widely used in LVDMs, utilizing the stable SD2.1-VAE version \cite{zhao2024cv}. SVD-VAE \cite{blattmann2023stable} enhances video VAE with an improved decoder, sharing the encoder with SD-VAE.
(2) TATS, CV-VAE, and OPS-VAE: These are video VAEs designed for video generation. TATS \cite{ge2022long} is a 3D video VQ-VAE for autoregressive video generation. CV-VAE \cite{zhao2024cv} and OPS-VAE \cite{opensora} were developed concurrently with our research.
(3) OD-VAE and OpenSora-Plan: OD-VAE \cite{chen2024odvaeomnidimensionalvideocompressor} is a contemporaneous video VAE with spatial and temporal compression. OpenSora-Plan \cite{pku_yuan_lab_and_tuzhan_ai_etc_2024_10948109} is an LVDM model trained with OD-VAE, compared using the Open-Sora Plan v1.2.0, 93×480p model.
(4) CogVideoVAE and CogVideoX-5B: CogVideoVAE \cite{yang2024cogvideox} is a 3D Causal VAE from CogVideoX, applying spatial and temporal downsampling. CogVideoX-5B \cite{yang2024cogvideox} supports 720p video generation.
We additionally fine-tuned two DiT models, CogVideoX and OpenSora-Plan, using our MTC-VAE Encoder to encode latent variables and text pairs as input, and evaluated their performance during inference.

\vspace{0.25cm}
\noindent\textbf{ Evaluation Metrics and Datasets.}
% Experimental Setup
In our study, we utilize two widely recognized large open-domain video datasets: WebVid-10M \cite{bain2021frozen} and Panda-70M \cite{chen2024panda}. Each video within the validation sets is standardized into a clip of 34 frames, maintaining a resolution of $256\times 256$. We conducted fine-tuning on two versions of MTC-VAE: MTC-VAE(ch=4), derived from OD-VAE \cite{chen2024odvaeomnidimensionalvideocompressor}, and MTC-VAE(ch=16), adapted from the VAE of CogVideoX-5B \cite{yang2024cogvideox}.

% Evaluation Metrics
\noindent \textbf{Metrics.} For video reconstruction, we consider the video compression rate \((VCPR = \frac{T}{t} \cdot \frac{H}{h} \cdot \frac{W}{w})\) to indicate compression efficiency and network complexity, along with three quality metrics: peak signal-to-noise ratio (PSNR) \cite{hore2010image}, structural similarity index measure (SSIM) \cite{wang2004image} and Learned Perceptual Image Patch Similarity (LPIPS) \cite{zhang2018unreasonable}. We determine Keyframe prediction accuracy (Acc.) by dividing correctly predicted frames by the overall number of frames. For video generation, performance is evaluated by VBench \cite{huang2023vbench}, VCPR, Speedup, and Latency.

% Video Reconstruction Experiments
% \noindent To assess the video reconstruction performance, we initially compare our MTC-VAE variants with baseline models using standard quality metrics. Subsequently, we test our compression algorithm's effectiveness by adjusting the control parameter \(w\) from 1.0 to 2.5 in 0.5 increments, highlighting its capacity to manage the compression rate. Furthermore, we evaluate various methods for assigning video segment compression rates: random allocation, H.264 (MPEG-4)-based assignment, and PSNR-guided allocation.

% Video Generation Experiments
\vspace{0.25cm}
\noindent \textbf{DiT Fine-tuning.} For video generation evaluation (as shown in the pseudocode~\ref{alg:vae_dit}), we compare our method with two leading models, CogVideoX-5B \cite{yang2024cogvideox} and OpenSoraPlan \cite{pku_yuan_lab_and_tuzhan_ai_etc_2024_10948109}. We fine-tune two versions of our model based on these architectures. 
Our primary objective is to minimize a noise prediction loss defined as:
\begin{equation}
\begin{aligned}
\mathcal{L}_{\text{DiT}} = \mathcal{L}_{\epsilon}(\boldsymbol{Z}; \boldsymbol{Z_{prompt}}).
\end{aligned}
\end{equation}
DiT is fine-tuned using the MTC-VAE encoder for a total of 20,000 steps. The training scheme incorporates a learning rate of \(1 \times 10^{-5}\) along with a batch size of 4. We utilize the OpenSoraPlan dataset, maintaining a resolution of \(256 \times 256\) pixels as specified by the OpenSoraPlan model outlined in \cite{pku_yuan_lab_and_tuzhan_ai_etc_2024_10948109}. Our model is developed at a resolution of \(720p\), based on the \(720p\) version of the CogVideoX model \cite{yang2024cogvideox}. This process is executed on 4 NVIDIA A100 GPUs, each with 80GB, within the PyTorch framework \cite{paszke2019pytorch}.

\begin{figure}[h]
    \centering
    \includegraphics[width=1\linewidth]{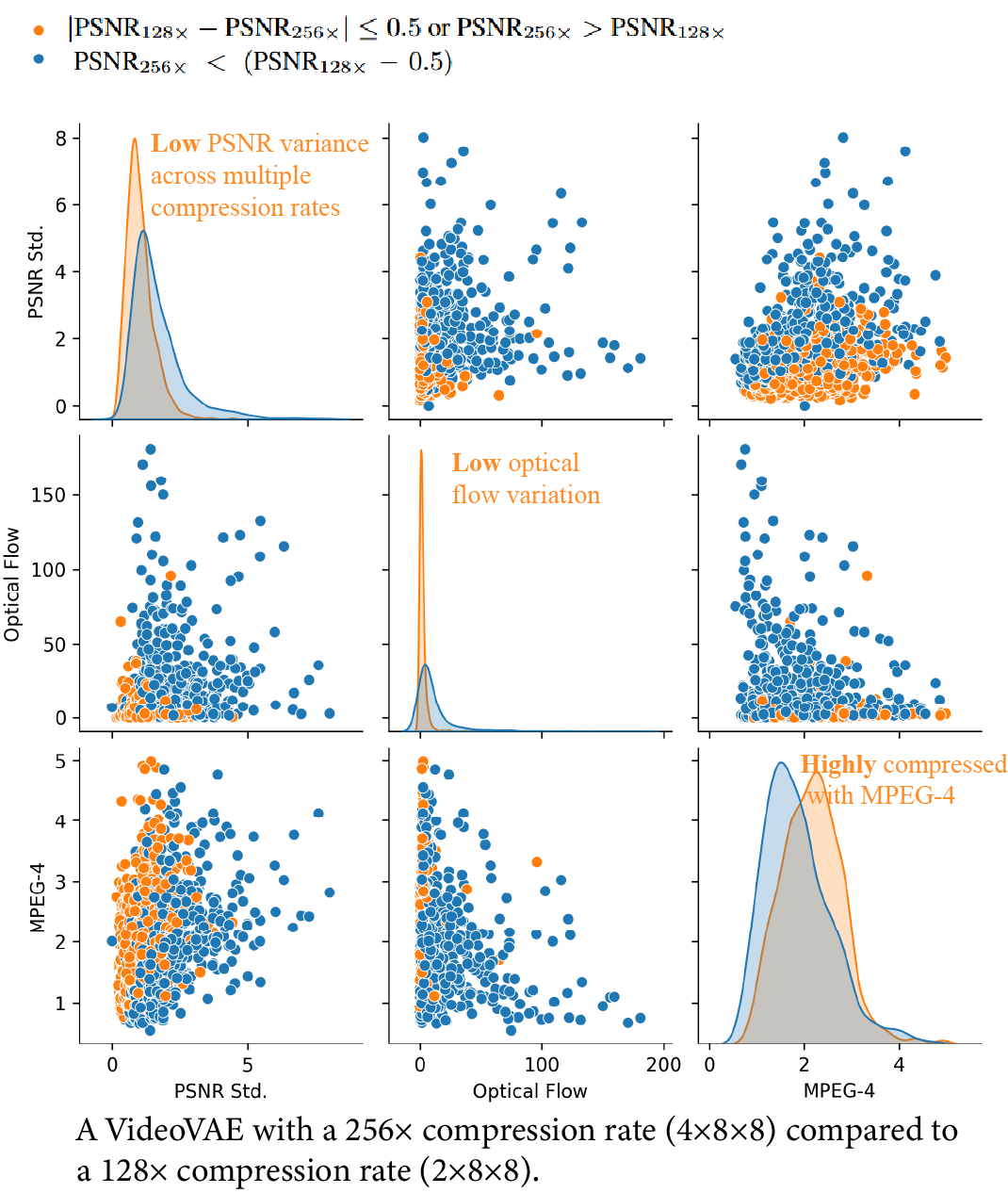}
    \caption{Scatter plot illustrates when the compression of 256×(4×8×8) provides equivalent or higher quality compared to 128×(2×8×8). Orange indicates cases where \(|\text{PSNR}_{128\times} - \text{PSNR}_{256\times}| \leq 0.5\) or \(\text{PSNR}_{256\times} > \text{PSNR}_{128\times}\) for a video segment, whereas blue signifies \(\text{PSNR}_{256\times} < (\text{PSNR}_{128\times} - 0.5)\). \textbf{PSNR Std.} denotes the standard deviation of PSNR values across compression levels 128×(2×8×8), 256×(4×8×8), 512×(8×8×8), and 1024×(16×8×8). \textbf{Optical Flow} refers to motion dynamics, and \textbf{MPEG-4} refers to the level of storage compression applied via the MPEG technique for the video. This data is gathered from 1,024 videos within the Pixabay collection, included in the OpenSoraPlan open-source dataset \cite{pku_yuan_lab_and_tuzhan_ai_etc_2024_10948109}.}
    \label{fig:scatter}
\end{figure}

\subsection{Discussion}
\label{sec:4.2}
\textbf{Which video segments can be temporally compressed higher?}
Figure \ref{fig:scatter} illustrates situations where videos can achieve comparable performance (\(|\text{PSNR}_{128\times} - \text{PSNR}_{256\times}| \leq 0.5\)), represented by the orange points. The distribution of orange and blue points demonstrates some overlap; however, they remain distinguishably apart. By analyzing the mean differences shown in the plots along the main diagonal, it becomes clear that: 1) Video segments that are highly compressible under MPEG-4 \cite{MPEG4-2002} generally maintain quality even with considerable temporal compression. 2) Video sequences with low optical flow, indicating smooth motion, tend to be resilient to higher compression levels. 3) Similar PSNR values across compression rates suggest higher compression is possible without significant quality loss.

% \noindent \textbf{VideoVAE's compression capability is distinct from that of the MPEG algorithm.} Generally, there is a direct correlation between the level of MPEG compression and the effectiveness of VAE's reconstruction of MPEG-compressed videos. In fact, the VAE's enhanced proficiency with high-level features allows it to more efficiently compress specific informational aspects of a video, a task often difficult for conventional compression methods. Consequently, in the bottom left plot of Figure \ref{fig:scatter}, orange dots appear at lower MPEG values \((MPEG \le 1)\) and lower PSNR Std,  demonstrating that certain videos cannot be well compressed by the MPEG algorithm, yet VAE can successfully handle them.

\subsection{Main Results}
\label{sec:4.3}
\textbf{Video Reconstruction Comparison.}
In Table\ref{Table1}, we report the results of four evaluation metrics for two models on two datasets.
Our key observations are:
1) SVD-VAE outperforms SD-VAE in video compression performance. On WebViD, it achieves a PSNR of 31.15 compared to SD-VAE's 30.19.
2) CV-VAE's method improves temporal consistency by supervising the latent space, reducing LPIPS to 0.0803.
3) OPS-VAE and OD-VAE achieve higher performance at higher compression rates through causal encoder-decoder designs and effective structural designs. With a VCPR of 256 on Panda-70M, they reach PSNR values of 31.06 and 30.49 respectively, with higher SSIM values of 0.8969 and 0.8970.
4) CogVideoVAE opts for higher channel counts, enhancing latent representation capacity.
5) Compared to the above models, our MTC-VAE method achieves higher video compression rates with minimal performance loss. On the WebVid-10M validation set and Panda-70M, MTC-VAE(ch=4) achieves a 73.3\% and 45.3\% improvement in compression rate over commonly used models such as OD-VAE and CogVideoVAE. MTC-VAE(ch=16) shows even greater improvements under the default \(w\) settings.
Besides, with higher compression rates, our MTC-VAE variants perform competitively against OD-VAE and CogVideoVAE. On the Panda-70M validation set, MTC-VAE(ch=4) achieves higher PSNR and SSIM scores than both CV-VAE and OD-VAE. Similarly, on the WebVid-10M validation set, our MTC-VAE(ch=16) delivers a 92.4\% improvement in compression rate while maintaining performance comparable to the CogVideoVAE model.

\noindent\textbf{Video Generation Comparison.}
In Table \ref{tab:vbench}, we present the results of video generation involving the fine-tuned DiT along with our MTC-VAE and other baseline models. The findings in Table \ref{tab:vbench} represent our comprehensive evaluation of the VBench Video Generation Benchmark. With respect to OpenSora-Plan, the MTC-OSP model enhances the Video Compression Rate (VCPR) from 256 to 441.9 and increases the speed of generation by a factor of 1.91, while the VBench score shows actual improvement. For CogVideoX-5B, even though the MTC-CogVideo model experiences a slight reduction in the VBench score (from 80.88 to 80.70), it boosts the VCPR to 366.7 from 256 and decreases latency from 4.43 seconds per frame to 3.12 seconds. Therefore, we deduce that:
1) Our MCT-VAE approach is compatible with current DiT models for video generation.
2) LVDM can benefit from multi-level temporal compression rates by accelerating the video generation process.

\noindent\textbf{Visual Results.}
For our analysis and qualitative review, we utilized the same prompts as CogVideoX and the extended prompts from VBench, as depicted in Figure \ref{fig2}. Our findings revealed that after fine-tuning with our MTC-VAE, DiT delivers comparable image quality in terms of visual aesthetics. Moreover, DiT with our MTC-VAE achieves higher compression rates at specific points, such as when characters perform minor actions within a scene (as seen in lines 3 and 4), or when they start to reflect or look into the distance (as seen in lines 1 and 2), among others, enabling our method to generate longer video outputs. Additional video results can be found in the supplementary materials.

% \begin{figure}[tbp]
% \centering
% \hspace{-0.5em}
% \includegraphics[width=.48\textwidth]{ICCV2025-Author-Kit/mtcvae/cmp_vaes.png}
% \caption{PSNR Performance Across Varying Compression Rates for VAEs. As compression rates increase from 400 to 800, MTC-VAE maintains higher PSNR values than OD-VAE \cite{chen2024odvaeomnidimensionalvideocompressor} and CogVideoVAE \cite{yang2024cogvideox}.}
% \label{fig:moe_vae_vae}
% \end{figure}
\begin{table}[ht]
\centering
\resizebox{\linewidth}{!}{
\begin{tabular}{llclcclc}
\toprule
Model & $\mathcal{L}_{\text{flow}}$ & $w$ & VCPR $\uparrow$ & PSNR $\uparrow$ & LPIPS $\downarrow$ & Acc.(\%) $\uparrow$ \\
\midrule
OD-VAE & -- & -- & 256.0 & 30.69 & 0.0502 & -- \\
OD-VAE & -- & -- & 512.0 & 29.99 & 0.0565 & -- \\
\hline
MTC-VAE & \xmark & 1.0 & 348.0 & 30.62 & 0.0564 & 99.8 \\
MTC-VAE & \checkmark & 1.0 & 370.0 & 30.76 & 0.0540 & 99.9 \\
MTC-VAE & \checkmark & 1.5 & 454.5 & 30.72 & 0.0569 & 99.9 \\
MTC-VAE & \checkmark & 2.0 & 748.0 & 30.32 & 0.0584 & 99.8 \\
MTC-VAE & \checkmark & 2.5 & 827.0 & 30.14 & 0.0601 & 99.8 \\
\bottomrule
\end{tabular}
}
\caption{Our ablation study of the MTC-VAE w/o flow-guided consistency loss and compared with OD-VAE at 256 and 512 compression rate. We find our flow-guided consistency loss useful in holding the VAE performance.}
\label{table:ablation}
\end{table}

\subsection{Ablation Study}
\label{sec:4.4}

% \noindent\textbf{PSNR performance across compression rates for VAEs.}
% Using 1024 videos from the OpenSoraPlan dataset, we compare models at different compression rates, as shown in Table~\ref{table:ablation}. We adjust the parameter $w$ in MTC-VAE for rough compression control, while OD-VAE and CogVideoVAE are fine-tuned for 512x. Our results indicate that as the compression rate increases, MTC-VAE consistently maintains higher PSNR values than OD-VAE.

\noindent\textbf{Use \(w\) to control the temporal compression rate.}
In order to gain a better understanding of the effects of \(w\) in our temporal compression algorithm, we utilize videos from the OpenSoraPlan dataset to compare models across various compression rates by adjusting \(w\) from 1.0 to 2.5 in 0.5 increments, as illustrated in Table~\ref{table:ablation}. Our findings indicate that: 1) Increasing \(w\) from 1.0 to 2.5 leads to an elevated compression rate. Significantly, MTC-VAE attains a VCPR of up to 827 with a PSNR of 30.14, which surpasses the PSNR offered by OD-VAE at a 512x compression rate. 2) The inclusion of optical flow loss \(L_{flow}\) enhances VAE performance; with \(w=1\), the use of \(L_{flow}\) during stage 2 training can boost the PSNR, increasing it to 30.76.

\begin{table}[htbp]
    \centering
    % \footnotesize
    \resizebox{\linewidth}{!}{
    \begin{tabular}{lccccc}
        \toprule
        Method & Idx & $f_c$ & PSNR↑ & LPIPS↓ & Acc.↑ \\
        \midrule
        MTC-VAE &begin & \xmark & 24.31 & 0.2390 & 0.41 \\
        MTC-VAE &end & \checkmark & 32.49 & 0.0353 & 1.00 \\
        \rowcolor[gray]{0.9} MTC-VAE &begin & \checkmark & \textbf{32.90} & \textbf{0.0338} & 1.00 \\
        \bottomrule
    \end{tabular}
    }
    \caption{Ablation study on extra embedding for segment marking. Adding the embedding to the first index achieves the best performance, while removing it significantly reduces accuracy.}
    \label{extra_emb}
\end{table}

\noindent \textbf{Extra embedding \(f_c\)}.
Learnable parameter \(f_c\) (C×1×1×1) adds to latent segments to mark each segment area, so the keyframe predictor can divide segments more easily.
To analyze the effectiveness of this embedding and determine whether the first position of each segment is better, we set up two controls: 1) The first option does not use the \(f_c\) embedding but still regards the first frame of each segment as the keyframe (predictor uses keyframe to segment latents) 2) Adding the embedding to the last frame of each segment, and regarding the last frame as the keyframe. We randomly selected 256 videos with 256 resolution and 34 frames each for this experiment.
The ablation (Table \ref{extra_emb}) shows: 1. Without extra embedding, the accuracy significantly decreases.
2. Adding to the beginning is better than adding to the end of the segment, possibly because the first frame in causal VAE contains denser information and is globally visible.
% Generation quality uses compression rate perplexity metric. 

\begin{table}[htbp]
    \centering
    % \footnotesize
    \resizebox{\linewidth}{!}{
    \begin{tabular}{lccccc}
        \toprule
        Selection Method & PSNR↑ & LPIPS↓ & SSIM↑ & VCPR↑ & Time(s/it)↓ \\
        \midrule
        Original (4×8×8) & 30.77 & 0.064 & 0.864 & 256.00 & 1.13 \\
        \midrule
        Random & 29.82 & 0.077 & 0.843 & 598.13 & \textbf{1.03} \\
        H.264 & 30.51 & 0.070 & 0.852 & 421.23 & 1.07 \\
        \rowcolor[gray]{0.9} PSNR-based & \textbf{30.67} & \textbf{0.069} & \textbf{0.859} & 492.00 & 1.22 \\
        \bottomrule
    \end{tabular}
    }
    \caption{Comparison of different quality function selection methods for our compression algorithm. The PSNR-based approach achieves the best reconstruction quality across all metrics with a modest increase in computational cost.}
    \label{tab:cmp_Q}
\end{table}

\noindent \textbf{The quality function of Compression Algorithm}.
In our compression algorithm, \(Q(·)\) is employed to assess video quality, as depicted in the scatter plot~\ref{fig:scatter} and our discussion. Although metrics like H.264 (MPEG-4), PSNR, or optical flow could be used, we observed significant overlaps in optical flow measurements, rendering it impractical for our needs. We evaluated three approaches—random selection, H.264 (MPEG-4) based selection, and PSNR-based selection—in terms of the quality-compression tradeoff for video sequences. When compared to analyzing the video compression ratio with H.264, we discovered that utilizing the PSNR of video reconstruction from VAE as \(Q(·)\) is more appropriate. This approach not only achieves higher PSNR and LPIPS but also enhances the VCPR while requiring a slightly greater compute budget.

\section{Conclusion}
In this work, we introduce MTC-VAE, a new approach that addresses the limitations of fixed compression rates by enabling multi-level temporal compression based on video content complexity. 
Our MTC-VAE demonstrates the ability to dynamically adjust the temporal compression ratio based on video content complexity, effectively balancing compression efficiency with high-fidelity reconstruction. The compression algorithm can adjust the compression rate by \(w\), allowing for fine-grained control over the temporal resolution. We analyse the video compression rate relationship with temporal compression rates, revealing improvements in both efficiency and performance.
We fine-tuned DiT with MTC-VAE, integrating our adaptive compression framework with diffusion transformers. This architecture optimizes the temporal distribution of information within the latent space, reducing information redundancy. Experiments and ablation studies on video reconstruction and generation demonstrate the effectiveness of our methods compared to fixed-rate compression approaches.
{
    \small
    \bibliographystyle{ieeenat_fullname}
    \bibliography{main}
}
\clearpage
\setcounter{page}{1}
\maketitlesupplementary
\nopagebreak

\section{Algorithm Description}

We present two key algorithms for our video compression and generation framework. Algorithm 1 details the MTC-VAE training process, while Algorithm 2 describes the VAE+DiT text-to-video generation pipeline.

% VAE训练流程和VAE+DiT推理流程伪代码
\begin{algorithm}
\caption{MTC-VAE Training Process}
\begin{algorithmic}
\REQUIRE Video $X$, compression rates $C = \{4, 8, 16\}$, weight $w$
\STATE Segment $X$ into $\{x_1, x_2, ..., x_N\}$ of equal length
\FORALL{$x_i \in \{x_1, x_2, ..., x_N\}$}
    \STATE $\boldsymbol{Q}(x_i) \leftarrow [Q(4,x_i), Q(8,x_i), Q(16,x_i)]$ \COMMENT{Quality vector}
    \STATE $\overline{Q}(x_i) \leftarrow \text{mean}(\boldsymbol{Q}(x_i))$
    \STATE $\sigma_Q(x_i) \leftarrow \text{std}(\boldsymbol{Q}(x_i))$
    \STATE $\alpha(x_i) \leftarrow \frac{1}{2} \cdot \frac{\overline{Q}(x_i) - Q_{\text{min}}}{Q_{\text{max}} - Q_{\text{min}}} + \frac{1}{2} \cdot (1 - \frac{\sigma_Q(x_i)}{\sigma_{Q_\text{max}}})$ \COMMENT{Adaptive weight}
\ENDFOR

\STATE $\{c_1, c_2, ..., c_N\} \leftarrow \arg\max\limits_{c_i \in C} \sum_{i=1}^N [(1-\alpha(x_i)) \cdot Q(c_i, x_i) + \alpha(x_i) \cdot \log_2(c_i) \cdot w]$ \COMMENT{Optimal rates}

\STATE \textbf{Encoder:}
\FORALL{$x_i$}
    \STATE $z_i \leftarrow \text{TemporalDownsample}(x_i^{\text{init}}, c_i)$ \COMMENT{Skip samplers based on $c_i$}
    \STATE $z_i^{1} \leftarrow z_i^{1} + f_{c_i}$ \COMMENT{Add compression embedding to first frame}
\ENDFOR

\STATE \textbf{Decoder:}
\FORALL{$z_i$}
    \STATE $p_i \leftarrow \text{KeyframePredictor}(z_i)$ \COMMENT{Keyframe probabilities}
    \STATE $\hat{z}_i \leftarrow \text{TemporalUpsample}(z_i, c_i)$ \COMMENT{Using ground truth $c_i$}
    \STATE $\hat{x}_i \leftarrow \text{Decode}(\hat{z}_i)$ \COMMENT{Reconstruct segment}
\ENDFOR

\STATE $\hat{X} \leftarrow \text{Concatenate}(\hat{x}_1, \hat{x}_2, ..., \hat{x}_N)$ \COMMENT{Final video output}
\RETURN $\hat{X}$
\label{alg:vae}
\end{algorithmic}
\end{algorithm}

\textbf{Algorithm 1: MTC-VAE Training Process.} This algorithm details our VAE training method for dynamic latent frame rates. We determine quality metrics for 4×, 8×, and 16× compression rates per video segment to derive an adaptive weight $\alpha(x_i)$, balancing reconstruction quality and efficiency. The encoder chooses optimal compression rates based on content complexity, embedding this rate data into the latent space. The decoder includes a keyframe predictor that identifies frames requiring higher precision.

\begin{algorithm}
\caption{VAE+DiT Text-to-Video Generation}
\begin{algorithmic}
\REQUIRE Text prompt $y$, noise schedule $\beta_1,...,\beta_T$, number of diffusion steps $T$, segment length $L$
\STATE $z_T \sim \mathcal{N}(0, I)$ \COMMENT{Sample initial noise}
\STATE $\hat{y} \leftarrow \text{TextEncoder}(y)$ \COMMENT{Encode text prompt}

\STATE \textbf{Diffusion Denoising:}
\FOR{$t = T$ \TO $1$}
    \STATE $\epsilon_\theta \leftarrow \text{DiT}(z_t, t, \hat{y})$ \COMMENT{Predict noise using DiT}
    \STATE $\mu_t \leftarrow \frac{1}{\sqrt{\alpha_t}}(z_t - \frac{\beta_t}{\sqrt{1-\bar{\alpha}_t}}\epsilon_\theta)$ \COMMENT{Predict mean}
    \IF{$t > 1$}
        \STATE $z_{t-1} \sim \mathcal{N}(\mu_t, \frac{\beta_t(1-\bar{\alpha}_{t-1})}{1-\bar{\alpha}_t}I)$ \COMMENT{Sample next state}
    \ELSE
        \STATE $z_0 \leftarrow \mu_1$ \COMMENT{Final denoised latent}
    \ENDIF
\ENDFOR

\STATE \textbf{VAE Decoding:}
\STATE $p \leftarrow \text{KeyframePredictor}(z_0)$ \COMMENT{Predict keyframe probabilities for each frame}
\STATE $K \leftarrow \{i \mid p_i > \text{threshold}\}$ \COMMENT{Identify keyframe positions}
\STATE Split $z_0$ into segments $\{z_1, z_2, ..., z_M\}$ based on keyframe positions $K$

\FORALL{segment $z_i$}
    \STATE $l_i \leftarrow \text{Length}(z_i)$ \COMMENT{Length of current latent segment}
    \STATE $c_i \leftarrow L / l_i$ \COMMENT{Calculate compression rate}
    \STATE $\hat{z}_i \leftarrow \text{TemporalUpsample}(z_i, c_i)$ \COMMENT{Upsample based on compression rate}
    \STATE $x_i \leftarrow \text{VAEDecode}(\hat{z}_i)$ \COMMENT{Decode latent to pixel space}
\ENDFOR

\STATE $X \leftarrow \text{Concatenate}(x_1, x_2, ..., x_M)$ \COMMENT{Merge segments into final video}
\RETURN $X$
\label{alg:vae_dit}
\end{algorithmic}
\end{algorithm}

\textbf{Algorithm 2: VAE+DiT Text-to-Video Generation.} This algorithm outlines our text-to-video generation process using a diffusion transformer (DiT) with our MTC-VAE framework. Starting with random noise and a text prompt, we perform iterative denoising using the DiT model conditioned on text embeddings. The denoised latent representation $z_0$ is then processed by our keyframe predictor to identify optimal segment boundaries. Each segment $z_i$ is upsampled according to its calculated compression rate $c_i$, which is determined by comparing the latent segment length to the target output length. Finally, the VAE decoder reconstructs each segment independently before concatenating them into the final video output.

\begin{figure}[h]
    \centering
    \includegraphics[width=1\linewidth]{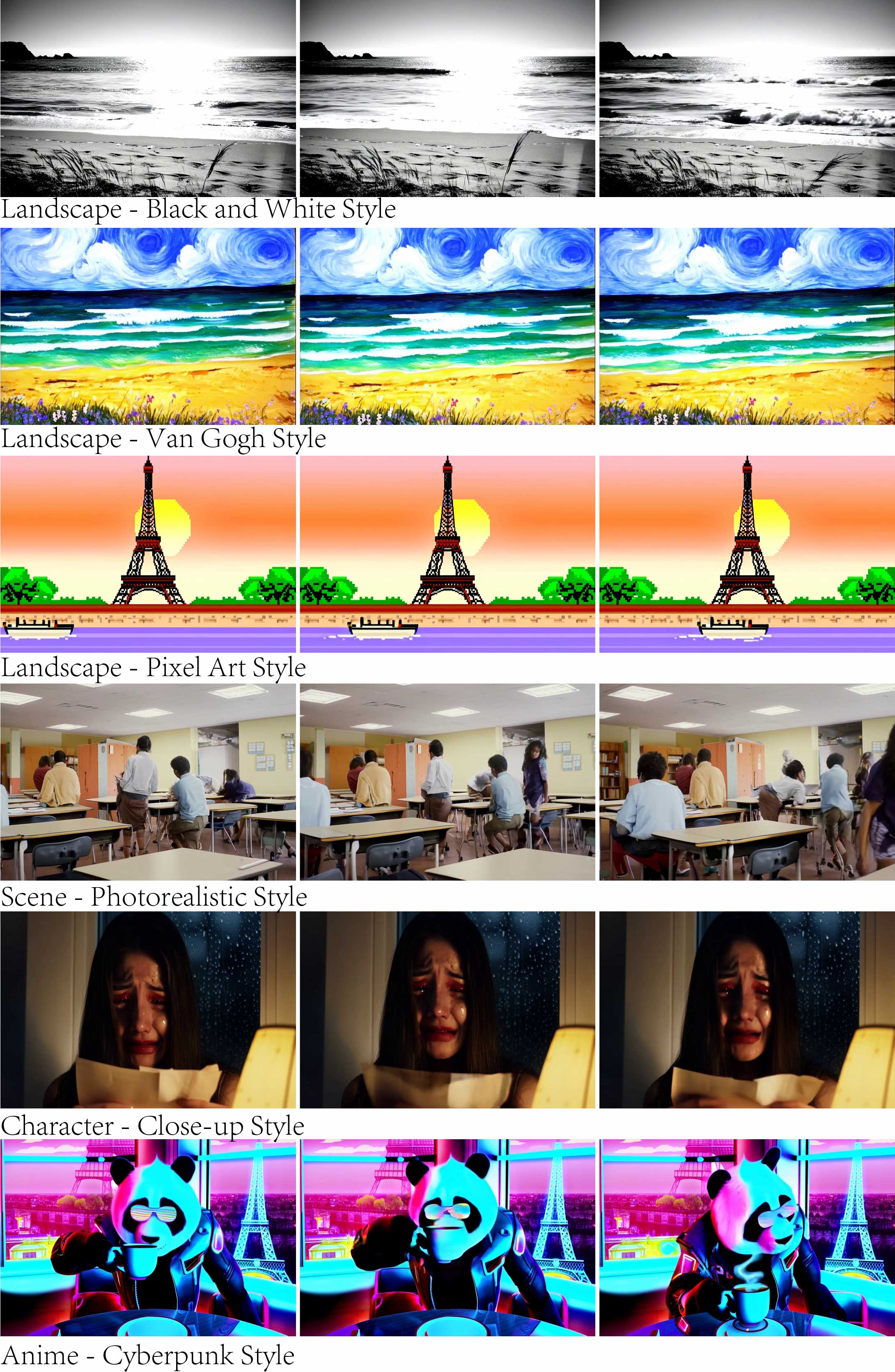}
    \caption{Visual comparison of our MTC-VAE model's performance across different scenes. Each row shows consistent style preservation despite varying compression rates, demonstrating robust visual quality maintenance across diverse content types.}
    \label{fig:appearance}
\end{figure}

% 对于风格的潜在影响
\section{Additional Results on VBench}

We conducted additional evaluations on the VBench benchmark to assess our model's capability in maintaining visual consistency across different compression rates. Figure~\ref{fig:appearance} demonstrates MTC-VAE's performance in preserving visual appearance while applying varying temporal compression.

As shown, our model successfully maintains consistent visual styles across diverse scenes - from black and white beach landscapes to stylized paintings with vibrant color palettes, photorealistic architectural structures, and character animations with artistic styling. These results confirm that our adaptive compression approach reduces computational requirements without compromising visual quality or style consistency, demonstrating robustness across VBench's diverse evaluation criteria.

% 可能的一些影响
\section{Impact Statement}
This paper presents work whose goal is to advance the field of Machine Learning, particularly in the domain of video generation. By introducing MTC-VAE (Multi-level Temporal Compression Variational Auto Encoder), we propose a simple to transfer framework that adjusts the latent frame rate based on video content complexity, while maintaining high reconstruction quality. This innovation has the potential to make video generation more efficient and scalable, enabling longer and higher-resolution video synthesis with reduced computational resources.

\subsection{Limitations and Future Work}
We acknowledge certain limitations in our current approach. For instance, we utilized three discrete compression rates to better leverage existing sampling parameters. Future work could explore fully trained VAE and DiT models using attention blocks to achieve arbitrary compression rates, enabling finer-grained compression. This would allow for more seamless integration between components, potentially enabling DiT to autonomously control video compression rates based on textual input.

While our keyframe predictor achieves high accuracy by framing compression rate prediction as a binary classification problem, prediction inaccuracies may still occur during generation. For frames with prediction confidence below 0.6, we default to the previous segment's compression rate (or 256× for the first segment). We attribute these limitations partly to insufficient pretraining and computational constraints, and encourage future work to address these challenges with more extensive training regimes.

\subsection{Broader Impact}
Our research contributes to making advanced video generation more accessible and environmentally sustainable. By reducing computational requirements, MTC-VAE enables researchers with limited resources to participate in cutting-edge video synthesis research. This democratization of technology could accelerate innovation across multiple sectors including creative arts, and scientific visualization.

We recognize that efficient video generation technologies carry dual-use potential. While they enable beneficial applications like improved educational content and medical simulations, they could also facilitate the creation of synthetic media for deception. We advocate for continued development of detection tools, watermarking techniques, and transparent disclosure practices to mitigate potential misuse. The computational efficiency of our approach also aligns with growing concerns about AI's environmental footprint, potentially reducing the carbon impact of video generation systems. We remain committed to responsible innovation that balances technological advancement with ethical considerations.

\end{document}